\documentclass{article}

\usepackage[preprint,nonatbib]{neurips_2023}

\usepackage[utf8]{inputenc} 
\usepackage[T1]{fontenc}    
\usepackage{hyperref}       
\usepackage{url}            
\usepackage{booktabs}       
\usepackage{amsfonts}       
\usepackage{nicefrac}       
\usepackage{microtype}      
\usepackage{xcolor}         
\usepackage{multirow}

\usepackage{graphicx}
\usepackage{wrapfig}
\usepackage{floatrow}
\usepackage{subcaption}
\usepackage[title,toc,titletoc,page]{appendix}

\hypersetup{
    colorlinks=true,
    linkcolor=red,
    citecolor=cyan,
    filecolor=magenta,      
    urlcolor=cyan,
    }

\usepackage{amsmath,amssymb}
\usepackage{textgreek}
\usepackage{algorithm}
\usepackage{algpseudocode}

\title{Any-to-Any Style Transfer:\\Making Picasso and Da Vinci Collaborate}

\author{Songhua Liu\quad Jingwen Ye\quad Xinchao Wang \\
 National University of Singapore \\
 \texttt{songhua.liu@u.nus.edu, \{jingweny,xinchao\}@nus.edu.sg}
}

\begin{document}

\maketitle

\renewcommand{\thefootnote}{\fnsymbol{footnote}}

\begin{abstract}
  Style transfer aims to render the style of a given image for style reference to another given image for content reference, and has been widely adopted in artistic generation and image editing. 
  Existing approaches either apply the holistic style of the style image in a global manner, or migrate local colors and textures of the style image to the content counterparts in a pre-defined way. 
  In either case, only one result can be generated for a specific pair of content and style images, which therefore lacks flexibility and 
  is hard to satisfy different users with different preferences. 
  We propose here a novel strategy termed \emph{Any-to-Any Style Transfer} to address this drawback, which enables users to interactively select styles of regions in the style image and apply them to the prescribed content regions. 
  In this way, personalizable style transfer is achieved through human-computer interaction. 
  At the heart of our approach lies in (1) a region segmentation module based on \emph{Segment Anything}, which supports region selection with only some clicks or drawing on images and thus takes user inputs conveniently and flexibly; (2) and an attention fusion module, which converts inputs from users to controlling signals for the style transfer model. 
  Experiments demonstrate the effectiveness for personalizable style transfer. 
  Notably, our approach performs in a plug-and-play manner portable to any style transfer method and enhance the controllablity. 
  Our code is available \href{https://github.com/Huage001/Transfer-Any-Style}{here}. 
\end{abstract}

\begin{figure}[!htbp]
\centering
  \includegraphics[width=0.8\textwidth]{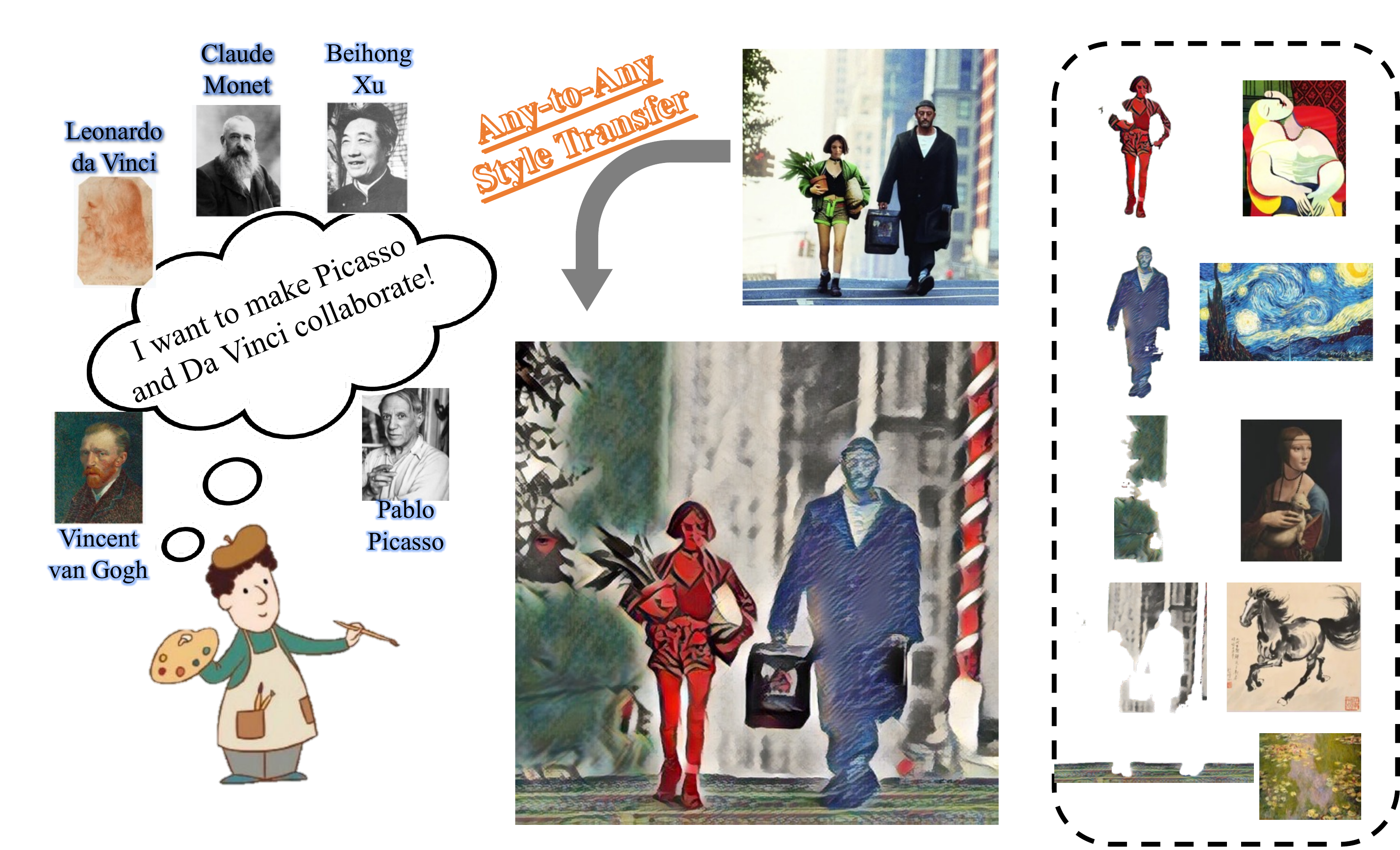}
  \caption{Any-to-any style transfer aims at supporting users with opportunities to customize styles for different areas of an image.}
  \label{fig:motivation}
\end{figure}

\section{Introduction}

Artificial intelligence generated content (AIGC), like GhatGPT~\cite{chatgpt,ouyang2022training}, DALL-E-2~\cite{ramesh2022hierarchical}, and Stable Diffusion~\cite{rombach2022high} has recently emerged as an exciting topic in the society, even beyond the artificial intelligence and computer science communities. 
Style transfer, as a popular task along this line, aims at applying style patterns like colors and textures of a reference image to a given content image, while preserving the semantic structure of the content~\cite{jing2019neural}. 
Such technique finds its applications in a broad spectrum of areas, including but not limited to image editing~\cite{karras2019style,patashnik2021styleclip,gal2022stylegan}, domain adaptation~\cite{kim2020learning,luo2020adversarial,fu2023meta}, and domain generalization~\cite{zhong2022adversarial,zhang2023adversarial,kim2023randomized,wang2021learning}. 

In the deep learning era, the pioneering work of Gatys \textit{et al.}~\cite{Gatys_2016_CVPR} introduces to minimize the joint content and style loss in the VGG feature space~\cite{simonyan2015very} in an iterative manner, yielding encouraging results but with a heavy computational overhead of online optimization for each content-style pair. 
To alleviate this issue, Johnson \textit{et al.}~\cite{johnson2016perceptual} and a series of following works~\cite{liu2017depth,wu2018direction,jing2018stroke,ulyanov2016texture,ulyanov2017improved,wang2017multimodal,lin2021drafting} propose to optimize a feed-forward network to accomplish style transfer. 
Once trained, the network can conduct style transfer for any input content image in real time. 
More flexibly, it is feasible for a single network to transfer styles of any style reference image to any content images during inference~\cite{huang2017arbitrary,chen2016fast,gu2018arbitrary,li2018learning,park2019arbitrary,jing2020dynamic,li2017universal,deng2020arbitrary1,ghiasi2017exploring,deng2020arbitrary2,yao2019attention,sheng2018avatar,wu2020efanet,shen2017meta,Liu_2021_ICCV,deng2022stytr2,wu2021styleformer,huo2021manifold,jing2022learning,LiuStyle2022ECCV,chen2021artistic,an2021artflow}, which are known as arbitrary style transfer methods and has become the main stream of research in style transfer. 

Specifically, most arbitrary style transfer methods work in the following way: (1) extract the content and style features by some backbone feature extractors; (2) calculate representations of content and style feature followed by some content-style interaction mechanisms to obtain the stylized feature; and (3) map the stylized feature to the image space by a feature decoder. 
Typically, in the 2nd step, existing approaches either adopt the global statistics to represent styles, or adopt attention mechanisms so that different locations in the content image may focus on different counterparts in the style image and thus improve local stylization effects. 
Although impressive results have been achieved, provided with a specific pair of content and style images, the trained network can only generate a single or unique result. 
Given that the performance of style transfer is largely subjective~\cite{wright2022artfid,yeh2018quantitative}, it is hard for previous methods to guarantee that the stylization results by a method would satisfy all users with different aesthetic preference. 

To alleviate the above issues, in this report, we explore a method enabling users to customize their own style transfer results, which is defined as \emph{Any-to-Any Style Transfer}, as shown in Fig.~\ref{fig:motivation}. 
In fact, one of our observations is that, whether a user likes a stylization result is largely dominated by what local style is applied for a specific content region. 
For instance, while rendering objects with Chinese water-ink style, determined by different objects they want to highlight, different users may prefer different ink intensities and they may want to apply the painting techniques of different regions. 
Along this line, we are motivated to explore a novel interaction, where users can control what content semantics apply what kind of styles conveniently. 
Given that semantic regions of an image is often ambiguous even for human, and that it requires much labour of users to mark target areas pixel-by-pixel, a core problem is then to find a user-friendly way to help users group pixels carrying consistent semantics for the content image and those constituting a local style element for the style image together. 

Recently, researchers from Meta AI Research published a new method for image segmentation termed \emph{Segment Anything Model}, or \emph{SAM}~\cite{kirillov2023segany}. 
Trained on an extensive dataset of 11 million images and 1.1 billion masks, it is a versatile algorithm that generates high-quality object masks based on input prompts, such as points or boxes. 
Thanks to the superior performance and flexibility to segment objects with SAM, we adopt it as a powerful tool to interact with users and help segment content and style images automatically. 
We have shown that with only a few clicks on the images, regions belonging to the same semantic concept or sharing similar colors and textures can be segmented in real time, which achieves simple yet effective human-computer interaction. 

Given a pair of masks of content and style images produced by interacting with SAM, which indicate which content regions should focus on which style regions, we devise an attention fusion module. 
It fuses the personalized control signal with the original content-style attention map, so that areas selected by users apply the selected style while those not selected by users still follow the default style transfer fashion. 
Experiments have demonstrated its effectiveness for any-to-any style transfer. 
Moreover, the proposed method works in a plug-and-play manner and is compatible with all existing style transfer methods to improve their controllability. 
In summary, the highlights of this report are as follows:
\begin{itemize}
    \item We propose a novel interaction fashion based on Segment Anything, which supports users to customize styles for different content components and thus achieves any-to-any style transfer;
    \item We devise an attention fusion module to take both users' inputs and model's inferences into consideration;
    \item The proposed method is portable to a variety of style transfer methods to make them personalizable. 
\end{itemize}

\begin{figure}
\centering
  \includegraphics[width=0.9\textwidth]{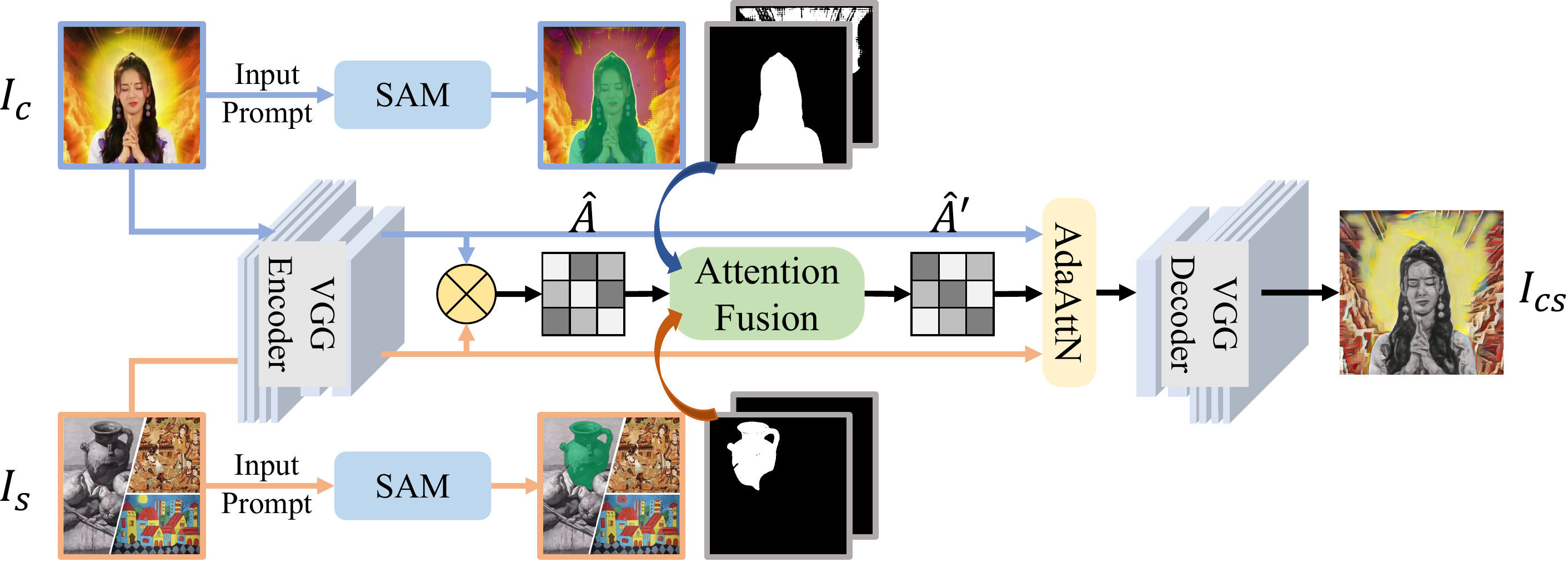}
  \caption{Overview of our any-to-any style transfer method. Users can specify which content regions apply styles of which style regions by interacting with Segment Anything Model.}
  \label{fig:pipeline}
\end{figure}

\section{Methods}

The primary goal of our any-to-any style transfer is to support users to specify which content regions should apply styles of which regions in the style image during style transfer. 
To this end, we devise the following pipeline: (1) content and style images are encoded by a pre-trained VGG-19 backbone and the default content-style attention map without personalization is computed; (2) Segment Anything Model (SAM) takes input prompts from users and obtains content and style segmentation masks; (3) The attention map in Step (1) is fused with the controlling signals from Step (2); and (4) the stylized feature is computed given the updated attention map and then decoded to the image space to derive the final stylization result. 
The overview of our method is shown in Fig.~\ref{fig:pipeline}. 
In this section, we first provide a brief introduction to the basic style transfer method adopted in this report. 
Then, we introduce how to get the content-style masks with SAM. 
And finally, we explain how to inject the controlling signals indicated by the masks to the original style transfer pipeline. 

\subsection{Preliminary}

Given a content image $I_c\in\mathbb{R}^{H_c\times W_c\times 3}$ and a style image $I_s\in\mathbb{R}^{H_s\times W_s\times 3}$, style transfer aims to generate a new image $I_{cs}\in\mathbb{R}^{H_c\times W_c\times 3}$, which reflects style patterns of $I_s$ while preserving content structure of $I_c$ simultaneously. 
In this report, we adopt \emph{AdaAttN}~\cite{Liu_2021_ICCV}, a simple yet effective attention-based style transfer baseline, as the basic method to build our any-to-any style transfer pipeline. 
It firstly adopt a VGG-19 encoder~\cite{simonyan2015very} pre-trained on the ImageNet dataset~\cite{deng2009imagenet}, denoted as $\Phi_{enc}$, to extract the content and style features $F_c\in\mathbb{R}^{h_cw_c\times f}$ and $F_s\in\mathbb{R}^{h_sw_s\times f}$:
\begin{equation}
    F_c=\Phi_{enc}(I_c),\quad F_s=\Phi_{enc}(I_s). 
\end{equation}
Three $1\times1$ convolution kernels, denoted as $g_q$, $g_k$, and $g_v$, are applied to derive the query, key, and value for attention computation:
\begin{equation}
    Q=g_q({\rm IN}(F_c)),\quad K=g_k({\rm IN}(F_s)),\quad V=g_v(F_s),\quad \hat{A}=Q\otimes K^\top,\quad A={\rm softmax}(\hat{A}),\label{eq:attn}
\end{equation}
where ${\rm IN}$ denotes the instance normalization operation and $\otimes$ represents matrix multiplication. 
AdaAttN aims to align the attention-weighted statistics, \textit{i.e.}, mean and standard deviation, of the target style, which are calculated by:
\begin{equation}
    S=A\otimes V,\quad V=\sqrt{A\otimes(V\cdot V)-M\cdot M},\quad F_{cs}=S\cdot {\rm IN}(F_c)+M,
\end{equation}
where $\cdot$ denotes element-wise multiplication. 
Finally, the stylized feature $F_{cs}$ is processed by a decoder $\Phi_{dec}$ symmetric to the encoder $\Phi_{enc}$:
\begin{equation}
    I_{cs}=\Phi_{dec}(F_{cs}). 
\end{equation}

\subsection{Interaction with SAM}

The Segment Anything Model (SAM) takes a set of points $P=\{(x_i,y_i,l_i)\}_{i=1}^{|P|}$ and / or a bounding box $b=(x_{lt},y_{lt},x_{rb},y_{rb})$ with an image $I$ as input and outputs a mask fulfilling the constraints indicated by the input prompts, where $(x_i,y_i)$ is the coordinates, $l_i$ is $0$ or $1$ indicating whether a point is belonging to the object or not, and $(x_{lt},y_{lt})$ and $(x_{rb},y_{rb})$ denote coordinates of the left-top corner and the right-bottom corner of the bounding box. 
The procedure of obtaining a mask $\Gamma$ can be written as:
\begin{equation}
    \Gamma={\rm SAM}(I,P,b). 
\end{equation}
In practice, the process of adding points and a bounding box works online: users can receive the current segmentation results in real time upon each point or bounding box is added. 
Thus, if SAM returns a mask including areas not expected by users, users can add more background points. 
Conversely, if a mask does not include enough areas expected by users, users can also add more foreground points. 
Masks of content and style images are selected in pair one-by-one, indicating which specific content regions should apply the style of which style regions. 
The final result of interaction with SAM are denoted as $\mathbf{\Gamma}=\{(\Gamma_{i,c},\Gamma_{i,s})\}_{i=1}^{|\mathbf{\Gamma}|}$. 

\subsection{Attention Fusion}

Given the mask pairs $\mathbf{\Gamma}=\{(\Gamma_{i,c},\Gamma_{i,s})\}_{i=1}^{|\mathbf{\Gamma}|}$ and the default attention map $\hat{A}$ in Eq.~\ref{eq:attn}, the proposed attention fusion is expected to modify the default attention map so that the final style transfer result would follow the control signals from users. 
The idea is to first down-sample the content and style masks to the same spatial resolution as $Q$ and $K$ in Eq.~\ref{eq:attn} respectively. 
Denote the down-sampled masks as $\tilde{\Gamma_{i,c}}$ and $\tilde{\Gamma_{i,s}}$, and we then set the rows of $\hat{A}$ masked by $\tilde{\Gamma_{i,c}}$ and the columns of $\hat{A}$ \emph{not} masked by $\tilde{\Gamma_{i,s}}$ to $-\infty$:
\begin{equation}
    \hat{A}[\tilde{\Gamma_{i,c}},\overline{\tilde{\Gamma_{i,s}}}]=-\infty,\quad 1\leq i\leq |\mathbf{\Gamma}|,\label{eq:edit}
\end{equation}
where the bar denotes the mask flipping operation. 
In this way, the attention map after ${\rm softmax}$ is edited so that the content areas selected by users would only focus on the style areas selected by users, with other entries set as $0$. 
Note that the operation in Eq.~\ref{eq:edit} performs in sequential over all mask pairs, which means if a latter content mask has some overlaps with some former ones, it would overwrite previous control signals. 

\section{Experiments}

\begin{figure}
\centering
  \includegraphics[width=\textwidth]{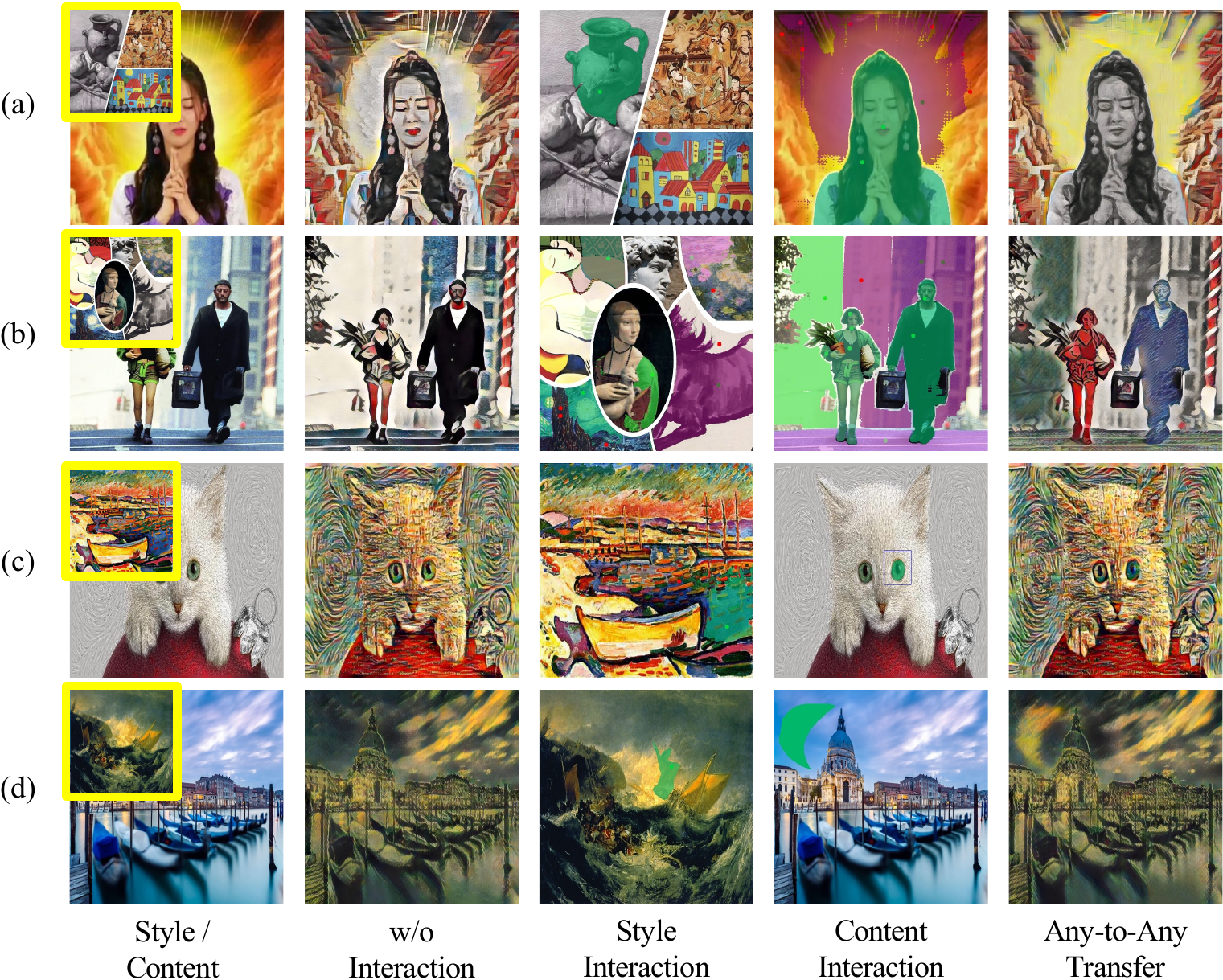}
  \caption{Results of our any-to-any style transfer. Users can select a region by putting foreground and background points (green and red points), draw a bounding box (blue rectangle), and / or draw a contour. After selection, the selected content regions would follow the styles of the corresponding style regions.}
  \label{fig:exp}
\end{figure}

Here, we show some example of our any-to-any style transfer in Fig.~\ref{fig:exp}. 
In Fig.~\ref{fig:exp}(a), our method can help users segment foreground and background within only one click so that they can apply different styles. 
Fig.~\ref{fig:exp}(b) is a more complex case where multiple masks are selected by users, which demonstrates the robustness of our method when handling complex scenes and multiple mask inputs. 
Fig.~\ref{fig:exp}(c) shows an example of selecting regions by a bounding box so that the mask is guided to fall in the box. 
Moreover, our method can also support users to draw some contours on the images so that regions in the contours would apply the selected style, as shown in Fig.~\ref{fig:exp}(d). 

\section{Related Works}

\subsection{Style Transfer}

Here, we present some related works of arbitrary style transfer and explain how our method for any-to-any style transfer is plugged into them to increase their controllability. 

\textbf{Local Transformation Based Style Transfer:} 
In local transformation based style transfer techniques, different spatial locations in the content image would apply different transformation functions. 
There are patch-similarity based solutions~\cite{chen2016fast,gu2018arbitrary,sheng2018avatar} and attention based methods~\cite{park2019arbitrary,yao2019attention,deng2020arbitrary1,Liu_2021_ICCV,deng2022stytr2,wu2021styleformer,LiuStyle2022ECCV}. 
Typically, these methods involve the computation of similarity map or attention map, to store the relation between each content location and style location. 
A higher score means this content position should pay more attention to this style position. 
For these methods, the attention fusion mechanism can be directly applied to modify the computed relation scores to achieve any-to-any style transfer. 

\textbf{Global Transformation Based Style Transfer:} 
A lot of works achieve arbitrary style transfer via global feature transformation, \textit{e.g.}, WCT~\cite{li2017universal}, AdaIN~\cite{huang2017arbitrary}, Linear style transfer~\cite{li2018learning}, DIN~\cite{jing2020dynamic}, MCCNet~\cite{deng2020arbitrary2}, and MAST~\cite{huo2021manifold}. 
In general, they can achieve the most attractable style transfer speed but dismiss stylized effects for local details a lot. 
To make these methods applicable for any-to-any style transfer, we need to create a pair of key and value for each style mask, including the default one. 
The value is the transformation function computed only through the masked areas and the default value is computed through the whole style image as usual. 
The attention map is modified so that content areas selected by users should focus on the keys with respect to the corresponding style areas. 

\textbf{Diffusion Based Style Transfer:} 
Most recently, diffusion models yield impressive performance in generation problems~\cite{ho2020denoising,dhariwal2021diffusion,ramesh2022hierarchical,rombach2022high}. 
In the literature of style transfer, a series of diffusion based techniques are proposed~\cite{xu2023stylerdalle,huang2022diffstyler,jeong2023training,yang2023zero,zhang2022inversion}. 
Text-guided style transfer is also possible. 
In general, diffusion based style transfer can be viewed as injecting additional conditions to the original model, which is usually achieved by either classifier-based guidance~\cite{dhariwal2021diffusion,liu2023more} or classifier-free guidance~\cite{ho2022classifier,nichol2021glide}. 
To make these methods achieve any-to-any style transfer, in both cases, we can perform denoising under the condition of each style regions or style text respectively and aggregate them through content masks in each steps, so as to achieve the effect that different content regions can apply styles of different style regions. 

\subsection{Applications of SAM}
Segment Anything Model (SAM) provides a versatile solution to image segmentation, which is one of the most fundamental and challenging problems in computer vision. 
It can also facilitate a lot of downstream applications, including image inpainting~\cite{yu2023inpaint}, object detection~\cite{yu2023h2rbox,wang2023segmenting}, and medical image processing~\cite{liu2023samm,deng2023segment,mohapatra2023brain}. 
Recent researches also focus on empirical studies of what SAM can and cannot do~\cite{ji2023sam,tang2023can,ji2023segment}. 
As a state-of-the-art foundation AI model, we expect that more applications of SAM in a wider spectrum of fields will emerged in the future.  

\section{Conclusion}

In this report, we propose a simple yet effective method based on Segment Anything Model (SAM) to achieve any-to-any style transfer, where users can control which content regions apply styles of which style regions flexibly and conveniently. 
Specifically, SAM firstly takes users' input prompts and generate pairs of content-style masks, which are then used to modify the default attention map to follow the input instructions from users and generate the final stylized feature for decoding. 
Experiments demonstrate the effectiveness. 
We also show that the proposed method is a plug-and-play component for all existing style transfer methods to improve their controllability. 

\medskip
{\small
\bibliographystyle{plain}
\bibliography{main}
}


\end{document}